\documentclass[conference]{IEEEtran}
\IEEEoverridecommandlockouts
\usepackage{cite}
\usepackage{amsmath,amssymb,amsfonts}
\usepackage{algorithmic}
\usepackage{graphicx}
\usepackage{textcomp}
\usepackage{xcolor}

\usepackage{footnote}
\usepackage[bottom]{footmisc}
\usepackage{threeparttable}
\usepackage{multirow}
\usepackage{tabularx}
\usepackage{ragged2e}
\usepackage{booktabs}
\usepackage{makecell}
\usepackage{xspace}
\usepackage{diagbox}
\usepackage{xcolor}
\usepackage{soul}
\usepackage[final]{pdfpages}
\usepackage{hyperref}
\usepackage{amsmath}
\usepackage{soul}
\newcommand{\mypara}[1]{\vspace{1ex}\noindent\textbf{#1} }

\def\BibTeX{{\rm B\kern-.05em{\sc i\kern-.025em b}\kern-.08em
    T\kern-.1667em\lower.7ex\hbox{E}\kern-.125emX}}
\begin{document}

\title{Road-Aware Anomaly Segmentation with Query-Guided Polygons and CLIP in Autonomous Driving
\thanks{ }
}

\author{Zhiran~Yan$^{1 \;*}$ and Gordon~Elger$^{1,2}$
\thanks{This work was supported by the Federal Ministry for Economic Affairs
and Climate Action in the project ``Gaia-X 4 AMS''.}
\thanks{$^{*}$
Corresponding author, email address: zhiran.yan@thi.de
    }
\thanks{$^{1}$
Institute of Innovative Mobility (IIMo), Technical University Ingolstadt
of Applied Sciences, Ingolstadt, 85049, Germany.
    }
\thanks{$^{2}$
Fraunhofer Institute for Transportation and Infrastructure Systems IVI, Ingolstadt, 85051, Germany.
        }%
}


\maketitle

\begin{abstract}
Traditional semantic segmentation models operate under a closed-set assumption and struggle to recognize unknown or unexpected objects—an essential capability for autonomous driving. As a result, such models often misclassify or overlook out-of-distribution (OOD) road anomalies, posing safety risks in open-world environments.
We present a lightweight, post-processing, road-aware anomaly segmentation framework that requires no retraining, no OOD data, and no auxiliary supervision. Our approach builds on a mask transformer–based segmentation network by exploiting query-level mask confidence and deriving a polygonal road prior to detect gap regions that may correspond to anomalies. To further suppress false positives, we introduce a CLIP-based zero-shot semantic filtering module using in-distribution prompts, with optional generalized OOD prompts.
By jointly leveraging spatial priors and semantic verification, our framework produces robust and interpretable anomaly predictions. Evaluation on three public benchmarks—Fishyscapes, SMIYC, and RoadAnomaly—shows consistently strong performance. In particular, our method outperforms the training-free baseline Maskomaly on most metrics and achieves the highest AP on Fishyscapes LostAndFound. These results demonstrate the practicality and deployability of our approach for real-world autonomous driving systems. Code is available at \url{https://github.com/chrisyan/RAAS}.
\end{abstract}



\section{Introduction}

Autonomous vehicles (AVs) are being deployed in increasingly complex environments, extending beyond constrained Operational Design Domains (ODDs) to city-scale and even inter-city driving~\cite{bogdoll2024anovox}. In such open-world settings, rare and unexpected events, such as fallen objects, debris, or unfamiliar road hazards, become more likely and pose critical safety risks. According to the German Insurance Association~\cite{dzaack2016comments}, collisions with unexpected obstacles remain a major cause of traffic accidents, highlighting the necessity of detecting out-of-distribution (OOD) objects that lie outside the training space of standard perception models.

While modern segmentation architectures excel in closed-set tasks such as semantic, instance, and panoptic segmentation~\cite{cheng2021mask2former}, they remain fundamentally restricted by predefined label spaces. Consequently, they often misclassify or entirely overlook unseen or unusual objects, creating blind spots in AV perception pipelines. 

\begin{figure}[t!]
\centering
\vspace{1em}
\includegraphics[trim=0 0 0 0,clip,width=1\linewidth]{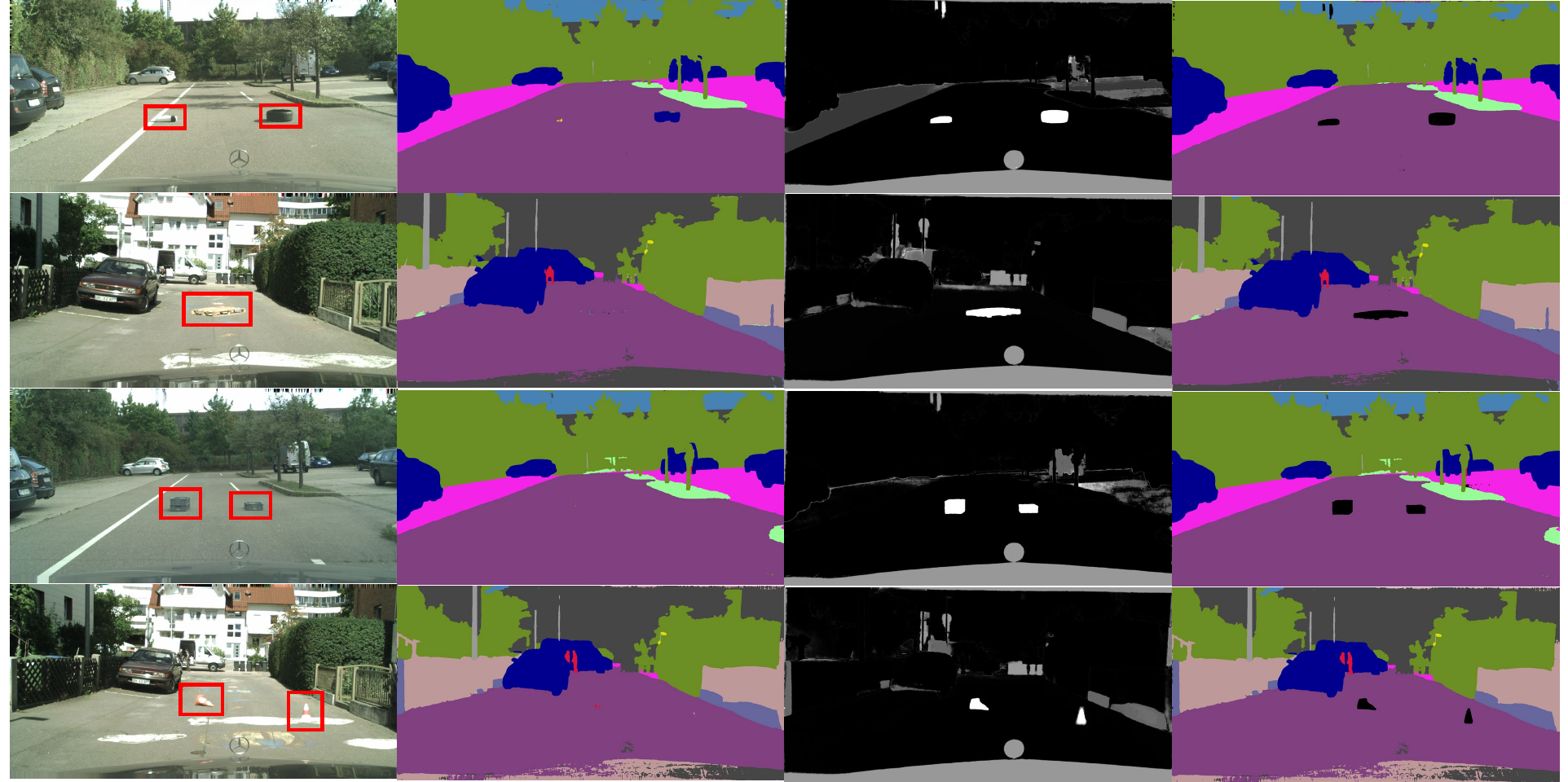}
\vspace{-5mm}
\caption{
Visual comparison between closed-set semantic segmentation and anomaly segmentation. From left to right: input RGB image with unknown obstacles (in red rectangle), semantic segmentation mask from a closed-set semantic segmentation model (which fails to detect out-of-distribution regions), anomaly map generated by our method (brighter pixels indicate higher anomaly scores), and anomaly map overlaid on the semantic segmentation mask. In the overlay, black pixels indicate detected anomalies, while colored regions correspond to known classes. Our method effectively highlights unexpected obstacles missed by closed-set models.
}
\vspace{0cm}
\label{fig:teaser}
\end{figure}

Camera-based anomaly segmentation aims to address this gap by identifying OOD regions at the pixel level. Existing approaches, including confidence scoring, reconstruction-based methods, and generative models~\cite{bogdoll2024anovox}, have demonstrated promising results but typically require retraining, outlier exposure (OE)~\cite{chan2021entropy}, or auxiliary networks, limiting their scalability and practicality. Recent post-processing approaches such as Maskomaly~\cite{ackermann2023maskomaly} provide a compelling alternative by exploiting query-level outputs from Mask2Former~\cite{cheng2021mask2former} at inference time; however, they still struggle to capture small or context-dependent anomalies due to the absence of explicit spatial or semantic filtering (see Fig.~\ref{fig:teaser}).

In this work, we propose a lightweight, plug-and-play anomaly segmentation framework designed specifically for autonomous driving. Our method runs entirely at inference time, requiring neither retraining nor OOD supervision. The framework consists of two complementary components:

\begin{itemize}
    \item \textbf{Road-aware anomaly proposal.}  
    We derive a polygonal road prior from road-related Mask2Former queries and identify gap regions along this polygon as anomaly proposals on the drivable surface.
    \item \textbf{CLIP-based semantic verification.}  
    Each candidate patch is evaluated using CLIP with natural-language prompts corresponding to Cityscapes~\cite{cordts2016cityscapes} classes. Optional generalized OOD prompts enhance robustness, enabling suppression of semantically valid ID objects and retention of true anomalies.
\end{itemize}

The refined proposals are then fused with a soft anomaly map obtained from query-level activations, yielding a unified anomaly score that combines spatial priors, semantic cues, and uncertainty-aware evidence.

\vspace{0.5em}
\noindent\textbf{Contributions.}  
Our main contributions are as follows:

\begin{itemize}
    \item \textbf{A road-aware, training-free anomaly segmentation framework.}  
    We introduce a lightweight post-processing pipeline that does not require retraining, OE, or auxiliary supervision. By adding spatial reasoning and semantic verification to query-level soft masks, our method substantially improves anomaly localization on the road surface.
    
    \item \textbf{A CLIP-based zero-shot semantic filtering module.}  
    We propose a patch-level CLIP classifier with both ID and optional generalized OOD prompts, enabling class-agnostic anomaly verification and effectively suppressing semantically plausible false positives.
    
    \item \textbf{Strong generalization across multiple benchmarks.}  
    Our method achieves the best AP on Fishyscapes LostAndFound (FS-LaF) and delivers competitive results on FS-Static, SMIYC and RoadAnomaly, consistently outperforming the baseline Maskomaly and rivaling more complex OE-based or retraining-heavy methods.
\end{itemize}

\section{Related Work}
\label{sec:related}

Out-of-distribution (OOD) detection for semantic segmentation has become a critical research topic, particularly for safety-critical applications such as autonomous driving. Following the taxonomy proposed by Bogdoll et al.~\cite{bogdoll2022anomaly}, existing detection approaches can be grouped into five main paradigms: \textit{confidence-based scoring}, \textit{reconstruction-based detection}, \textit{generative modeling}, \textit{feature-based extraction}, and \textit{prediction-based reasoning}.

\subsection{Confidence-Based Methods}

Confidence-based approaches estimate uncertainty directly from the outputs of segmentation networks. Some methods do not require retraining or auxiliary models, instead relying solely on model outputs such as softmax probabilities or logits. A foundational example is the Maximum Softmax Probability (MSP) proposed by Hendrycks et al.\cite{hendrycks2016baseline}, which detects misclassified and out-of-distribution (OOD) samples using the maximum softmax score. This idea has been extended by logit-based methods, such as MaxLogit\cite{hendrycks2019scaling} and Standardized Max Logits (SML)~\cite{jung2021standardized}, which also operate without additional training.

In contrast, more recent approaches such as Meta-OOD~\cite{chan2021entropy} and PEBAL~\cite{tian2022pixel} incorporate outlier exposure and require additional training stages. Meta-OOD adds a confidence estimation branch to the segmentation network, which is jointly trained using an entropy-based margin loss. PEBAL further retrains the segmentation model using an energy-bounded learning objective that explicitly separates ID and OOD pixels. These methods achieve improved performance at the cost of increased training complexity and reliance on auxiliary OOD data.

\subsection{Reconstruction-Based Methods}

Reconstruction-based methods assume that regions containing OOD content are harder to reconstruct than in-distribution areas. Image Resynthesis~\cite{lis2019detecting} and SynBoost~\cite{di2021pixel} use generative models to reconstruct the entire image and measure pixel-wise reconstruction error. JSR-Net~\cite{vojir2021road} explicitly reconstructs drivable road surfaces and flags mismatched regions as anomalies. While effective, these approaches tend to be computationally expensive and less suitable for real-time deployment.

\subsection{Generative Modeling}

Generative approaches aim to model the distribution of inlier data and detect anomalies as deviations from this distribution. For instance, DenseHybrid~\cite{grcic2022densehybrid} combines discriminative and generative signals using shared feature representations. Diffusion-based models like DOOD~\cite{galesso2024diffusion} employ score-matching to compute pixel-wise anomaly likelihoods, achieving strong generalization across datasets like ADE-OOD.

\subsection{Feature-Based Distance Estimation}

These methods analyze the internal feature representations of segmentation networks to detect OOD samples based on distance metrics. A representative example is the Mahalanobis distance~\cite{lee2018simple}, which computes class-conditional deviations in feature space under the assumption of Gaussianity. Other approaches extract uncertainty from intermediate features or use embedding clustering for instance separation. Notably, U3HS~\cite{cap2023u3hs} identifies high-uncertainty regions and clusters their embeddings into unknown object instances, enabling holistic segmentation without prior knowledge of unknowns. These approaches often require storage of class prototypes or training data statistics, introducing memory and tuning overhead.

\subsection{Prediction-Based Reasoning}

Prediction-based methods attempt to infer future frames or reconstruct masked content under the assumption of normality. Anomalies are then identified as regions where the prediction diverges significantly from observation. While this approach has been widely explored in video anomaly detection and action recognition, it remains relatively underutilized in the context of semantic segmentation. Recent work by Tian et al.~\cite{tian2024latency} highlights this gap by introducing a large-scale photorealistic video anomaly segmentation dataset along with temporally informed metrics, such as \textit{temporal consistency} and \textit{latency-aware streaming accuracy}. Although their work does not introduce a prediction-based method, it provides a benchmarking framework that emphasizes the importance of temporal reasoning in road anomaly detection. This shift toward video-based evaluation is expected to drive the development of future prediction-informed anomaly segmentation methods.

\subsection{Outlier Exposure and OOD Supervision}

Outlier Exposure (OE)~\cite{chan2021entropy} is a common strategy for enhancing pixel-wise OOD detection by pasting object regions from auxiliary datasets (e.g., COCO, ADE20K) onto inlier scenes such as Cityscapes, creating synthetic anomalies for supervision. Meta-OOD\cite{chan2021entropy} maximizes entropy over these pasted regions to encourage model uncertainty on unknown classes. PEBAL\cite{tian2022pixel} introduces an abstention class and uses OE to assign high energy to OOD pixels through an energy-biased loss. RPL\cite{liu2023residual} leverages OE to train a residual learner that captures anomaly-specific deviations from standard features. DenseHybrid\cite{grcic2022densehybrid} combines OE with both likelihood and softmax-based confidence for joint generative-discriminative OOD scoring. RbA~\cite{nayal2023rba} applies OE in a region-based framework by rejecting masks unrecognized by any known class, improving region consistency and reducing overconfidence.

While effective, these methods rely on additional training with synthetic OOD data, which limits their practicality in real-time or plug-and-play systems. In contrast, our approach is training-free and operates entirely at inference time, making it readily deployable in existing semantic segmentation pipelines.

\subsection{Mask-Based and Instance-Aware Anomaly Segmentation}
Recent research has increasingly shifted from per-pixel classification toward instance-aware or mask-based reasoning in anomaly segmentation, aiming to better capture uncertainty and reduce false positives.  
Mask2Anomaly~\cite{rai2023mask2anomaly} introduces a unified mask-classification architecture that simultaneously addresses anomaly segmentation, open-set semantic segmentation, and panoptic segmentation. It employs global masked attention, mask contrastive learning, and mask-level refinement to improve the separation of unknown instances. The anomaly score in Mask2Anomaly is derived using the Maximum Softmax Probability (MSP), but rather than computing softmax on a pixel-by-pixel basis, it calculates mask-level softmax scores, multiplies them with the sigmoid-activated mask logits, and then marginalizes the result.
RbA~\cite{nayal2023rba} aggregates region-level mask queries to estimate anomaly likelihood, leveraging instance-based representations for decision-making. Maskomaly~\cite{ackermann2023maskomaly}, built on top of Mask2Former~\cite{cheng2021mask2former}, post-processes the model’s multi-mask output at inference time to generate soft anomaly maps, using mask confidence as a proxy for anomaly likelihood.
These methods mark a significant difference from traditional pixel-wise confidence estimation, shifting toward structured, object-level anomaly discovery. By leveraging instance-aware embeddings and mask-level reasoning, they achieve improved detection granularity and robustness, especially in complex real-world scenes containing multiple unknown objects.

Our method follows this instance-aware inference paradigm. Like Maskomaly, it operates entirely at inference time, requiring neither retraining nor access to external OOD data. However, we go a step further by introducing road-aware query mask selection and CLIP-guided  \cite{radford2021learning} semantic filtering. 

\subsection{Zero-Shot 2D Classification with CLIP}

CLIP (Contrastive Language-Image Pre-training)~\cite{radford2021learning} enables zero-shot image classification by aligning visual and textual representations in a shared embedding space. Trained on 400 million image-text pairs collected from the internet, CLIP uses a contrastive loss to maximize similarity between matched image-text pairs and minimize it for mismatches. It combines powerful vision backbones, such as ResNet-50 and ViT, with a transformer-based text encoder. In a zero-shot setting, CLIP computes embeddings for both images and text, evaluates their cosine similarities, and applies a softmax over these similarities to produce class probabilities. Performance is enhanced through prompt engineering—for example, using phrases like “a photo of a car on a street” instead of plain class names. Unlike traditional classifiers trained on closed sets, CLIP's flexibility in defining arbitrary class sets makes it particularly suitable for open-world tasks such as anomaly detection.

In this work, CLIP serves as a zero-shot 2D semantic verifier to refine initial anomaly candidates. Specifically, we extract image patches corresponding to initial road anomaly regions (obtained via mask-level uncertainty) and evaluate their semantic similarity against a predefined set of 19 in-distribution (ID) classes using textual prompts. If a patch exhibits low similarity to all ID prompts (e.g., “a car,” “a traffic light,” “a person”), it is confirmed as an anomaly. Otherwise, it is suppressed as a likely in-distribution object.

\section{Method}

\begin{figure*}[t]
\centering
\includegraphics[width=0.95\linewidth]{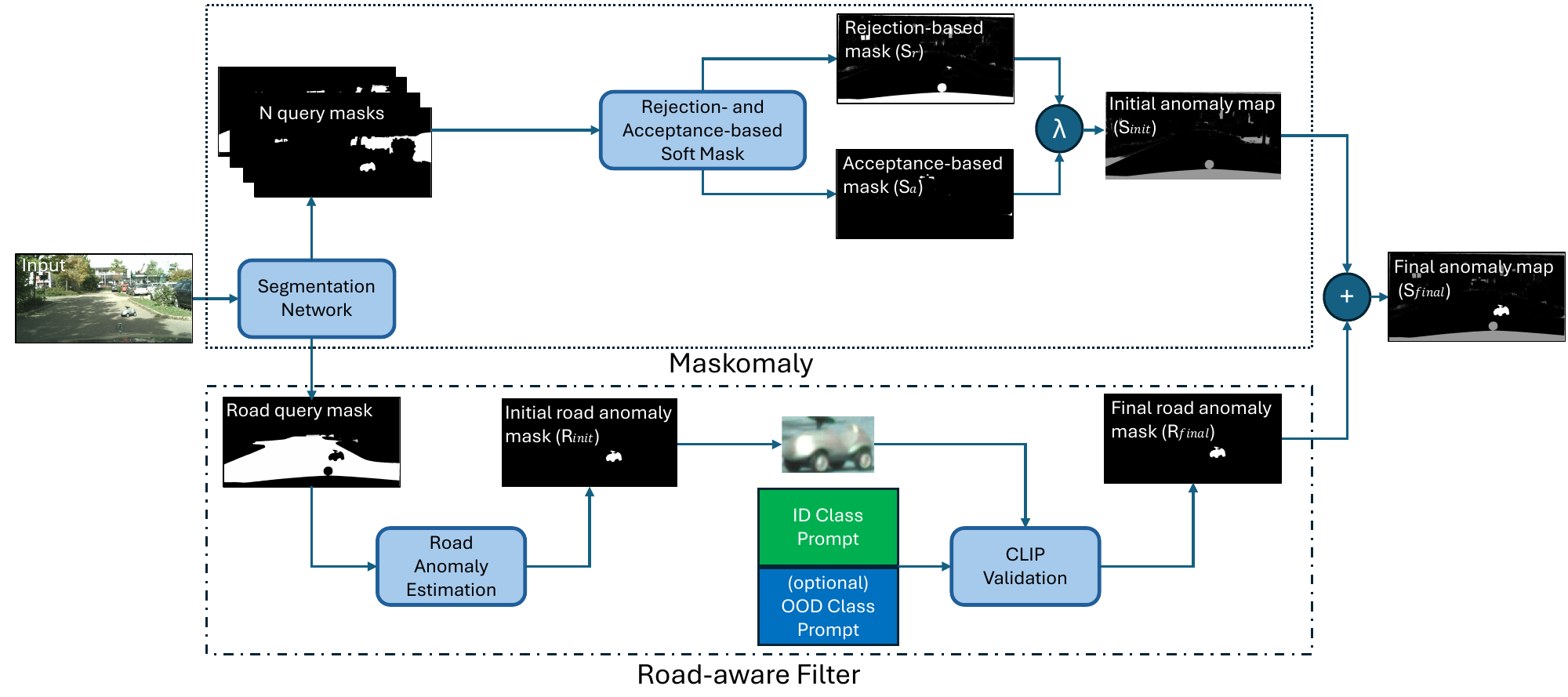}
\vspace{-2mm}
\caption{
Overview of the proposed road-aware anomaly segmentation pipeline, which comprises two branches: the baseline Maskomaly branch (in upper dashed box) and the proposed road-aware filtering module (in lower dashed box). The input image is processed by a segmentation network (Mask2Former), producing query masks and classification scores. Rejection- and acceptance-based soft masks are computed and fused to obtain an initial anomaly map. In parallel, a road-specific query mask extracts a filled road polygon, from which potential road anomalies are identified as gaps. These candidates are refined through CLIP-based semantic validation using ID-class prompts, optionally augmented with OOD prompts. Finally, the validated anomalies are integrated with the initial anomaly map, yielding a spatially focused and semantically filtered final anomaly map.
}

\label{fig:overview}
\end{figure*}

Our goal is to accurately segment anomalous regions in urban driving scenes using only in-distribution training data. Building on the Maskomaly framework~\cite{ackermann2023maskomaly}, we introduce a road-aware spatial filtering step that leverages semantic priors to improve anomaly segmentation performance, particularly within drivable areas. The proposed method adopts the transformer-based Mask2Former~\cite{cheng2021mask2former} as its segmentation backbone and operates entirely at inference time. As illustrated in Fig.~\ref{fig:overview}, the pipeline integrates three key modules: (i) a rejection- and acceptance-based soft mask inherited from Maskomaly, (ii) a road-query polygon mask for spatial filtering, and (iii) CLIP-based semantic validation.

\subsection{Rejection- and Acceptance-based Soft Mask}
\label{subsec:softmask}

Mask2Former takes a set of $N$ learnable queries (e.g., $N=100$) as input.  
Each query $i$ is decoded into a class-probability vector $\mathbf{c}_i \in \mathbb{R}^{K}$ and a pixel-wise mask probability map $\mathbf{m}_i \in [0, 1]^{H \times W}$.  
Building upon these outputs, Maskomaly~\cite{ackermann2023maskomaly} introduces two complementary soft masks—\textit{rejection-based} and \textit{acceptance-based}—to construct an initial anomaly map.  
The key idea is to suppress confidently recognized inlier regions while highlighting spatial areas that are semantically uncertain or atypical.

\begin{itemize}
    \item \textbf{Rejection-based soft mask ($S_r$)}:  
    This mask suppresses anomaly evidence in regions confidently explained by known in-distribution (ID) classes.  
    Initialized as an all-one map, $S_r$ is iteratively updated for each query $i$ whose most probable class is non-void and exceeds a confidence threshold $\tau_c$ (e.g., 0.7):
    \begin{equation}
        S_r \leftarrow \min\big(S_r,\ 1 - \mathbf{m}_i \cdot \max(\mathbf{c}_i)\big)
        \label{eq:rejection}
    \end{equation}
    Pixels covered by multiple high-confidence masks are consequently driven toward zero, effectively rejecting confidently classified inlier regions.  
    In practice, additional refinements such as ground suppression and boundary smoothing are applied to stabilize the rejection process.

    \item \textbf{Acceptance-based soft mask ($S_a$)}:  
    In contrast, $S_a$ aggregates regions potentially related to unknown or out-of-distribution (OOD) content.  
    Initialized as an all-zero map, it fuses a predefined subset of query masks empirically identified as responsive to anomalous structures.  
    For each such query $i \in \mathcal{A}$, where $\mathcal{A} = \{49,\,31,\,83,\,32\}$ in the original Maskomaly implementation, we update:
    \begin{equation}
        S_a \leftarrow \max\big(S_a,\ \mathbf{m}_i \cdot \max(\mathbf{c}_i)\big)
        \label{eq:acceptance}
    \end{equation}
    This aggregation highlights pixels that are frequently activated by anomaly-sensitive queries, capturing semantically inconsistent or structurally irregular regions.
\end{itemize}

The two masks encode complementary cues: $S_r$ provides \textit{negative evidence} by rejecting inlier predictions, whereas $S_a$ contributes \textit{positive evidence} from anomaly-related activations.  
The final anomaly probability map is obtained by a weighted combination of the two:
\begin{equation}
    S_\text{init} = \lambda\, S_r + (1 - \lambda)\, S_a, \qquad \lambda = 0.6
    \label{eq:fusion}
\end{equation}
This fusion balances confident inlier suppression and anomaly promotion, yielding a robust and fine-grained anomaly localization map during inference.

\subsection{Initial Road Anomaly Estimation Using Query Mask}
\label{subsec:roadmask}

To localize anomalies on the drivable surface, we exploit the observation that a specific query in Mask2Former consistently attends to the road region.  
Empirically, the 20\textsuperscript{th} query mask is found to correspond to the road in urban scenes.  
We threshold its probability map at $0.5$ and apply contour detection followed by polygon filling to construct a binary road polygon mask $P_{\text{road}}$.  
The initial road anomaly mask is then defined as the complementary region within the road polygon that is not covered by the road query prediction:
\begin{equation}
R_{\text{init}} = P_{\text{road}} \land \lnot M_{\text{road}},
\label{eq:init_ano}
\end{equation}
where $M_{\text{road}}$ denotes the binarized road query mask.  
This step provides a coarse estimation of anomalous areas that physically lie on the road surface but were not recognized as part of it.

\subsection{CLIP-Based Semantic Validation}
\label{subsec:clip}

While $R_{\text{init}}$ highlights potential road anomalies, it may still include false positives such as vehicles, pedestrians, or other known objects located within the road polygon.  
To refine these detections, we employ semantic patch validation using CLIP, a zero-shot vision–language model capable of classifying image regions based on natural-language prompts.  

Each connected component in $R_{\text{init}}$ is extracted as an RGB patch and classified with CLIP using a set of $19$ text prompts corresponding to the Cityscapes in-distribution (ID) classes (e.g., \textit{``a photo of a normal car on the road''}).  
In addition, we include several broad out-of-distribution (OOD) prompts, such as \textit{``something unusual in a driving scene''} or \textit{``an unexpected object on the road''}, to capture unforeseen anomalies without specifying explicit categories.  

For each patch, if the maximum CLIP confidence among ID prompts exceeds a threshold $\tau_{\text{ID}}$ ($0.8$), the region is classified as in-distribution and its anomaly score is set to a low value ($0.05$).  
Otherwise, the region is considered out-of-distribution and assigned a high anomaly score ($0.99$).  
This semantic filtering effectively removes contextually valid road entities while preserving truly anomalous objects.

\subsection{Final Anomaly Map}
\label{subsec:fusion}

The final anomaly map $S_{\text{final}}$ is obtained by fusing the soft anomaly map $S_{\text{init}}$ (Eq.~\ref{eq:fusion}) with the CLIP-validated road anomalies.  
For each patch validated as anomalous by CLIP, the corresponding pixels in $S_{\text{final}}$ are overwritten with a high anomaly confidence of $0.99$, ensuring semantic consistency between visual and linguistic reasoning.  
All other regions retain their original soft anomaly scores.  
Formally, the fusion rule is expressed as:
\begin{equation}
S_{\text{final}}(x, y) =
\begin{cases}
0.99, & \text{if } (x, y) \in R_{\text{final}}, \\
S(x, y), & \text{otherwise},
\end{cases}
\label{eq:final_ano}
\end{equation}
where $R_{\text{final}}$ denotes the set of pixels belonging to CLIP-validated anomalous patches.  
This integration step combines spatial priors from the road mask with semantic cues from CLIP, leading to improved precision and robustness in road-specific anomaly segmentation.


\section{Experiments}
\label{sec:usecases}

\subsection{Experimental Setup}
\mypara{Implementation Details.} In our experiments, we use the Mask2Former  \cite{cheng2021mask2former} segmentation model with a Swin-L \cite{liu2021Swin} backbone initially trained on the Cityscapes \cite{cordts2016cityscapes} dataset.
Cityscapes provides 2975 training, 500 validation, and 1525 test images at a resolution of $2048 \times 1024$, annotated with 19 in-distribution (ID) semantic classes.   All experiments are conducted on an {NVIDIA GeForce RTX~4080} GPU.  


\mypara{Datasets.}We evaluate our method on three real-world anomaly segmentation datasets containing driving-scene images with unknown road obstacles, demonstrating its generalization ability. Fishyscapes \cite{blum2019fishyscapes} consists of two subsets. FS-LaF contains 100 validation images ($2048 \times 1024$) from the LostAndFound dataset. FS-Static includes 30 validation images ($2048 \times 1024$) created by overlaying anomalous objects from Pascal VOC \cite{everingham2015pascal} onto Cityscapes images. Segment-Me-If-You-Can (SMIYC) \cite{chan2021segmentmeifyoucan} provides two complementary benchmarks: SMIYC-Anomaly with 10 validation and 100 test images ($1280 \times 720$) containing diverse anomalous objects.
SMIYC-Obstacle with 30 validation and 327 test images ($1920 \times 1080$) focusing on small road hazards under challenging conditions such as nighttime or adverse weather.
RoadAnomaly \cite{lis2019detecting} includes 60 images ($1280 \times 720$) depicting various unexpected road objects such as animals, rocks, and debris.
Across all datasets, we follow prior work \cite{lis2019detecting, di2021pixel, grcic2022densehybrid, hendrycks2016baseline, hendrycks2019scaling, jung2021standardized,chan2021entropy, tian2022pixel, liu2023residual, nayal2023rba, rai2023mask2anomaly, ackermann2023maskomaly} in defining out-of-distribution (OOD) objects as those not belonging to the 19 semantic classes of Cityscapes.


\mypara{Baselines.}We compare the proposed method against a curated set of state-of-the-art anomaly segmentation approaches, following the taxonomy proposed by Breitenstein et al.\cite{bogdoll2022anomaly}. The selected baselines span four major paradigms: confidence-based methods (e.g., MaxLogit\cite{hendrycks2019scaling}, SML~\cite{jung2021standardized}, Maskomaly~\cite{ackermann2023maskomaly}), reconstruction-based methods (e.g., SynBoost~\cite{di2021pixel}), generative modeling (e.g., DenseHybrid~\cite{grcic2022densehybrid}), and feature-based approaches (e.g., RPL+CoroCL~\cite{liu2023residual}).

Several of these methods require re-training the base segmentation model (Re-T), outlier exposure (OE), or re-training an external networks (EN). In contrast, our method is a lightweight, plug-and-play solution applied solely at inference time—requiring no re-training, no OOD supervision, and introducing no additional model components.

Notably, our method builds upon the Maskomaly pipeline, enhancing it with a spatial filtering module that leverages road semantics to suppress false positives near non-navigable regions. This results in improved localization of fine-grained and road-adjacent anomalies, particularly in complex urban scenarios. For all baselines, we adopt official implementations released by the respective authors whenever available to ensure a fair comparison.

\mypara{Evaluation Metrics.} We evaluate anomaly segmentation using three standard pixel-level metrics.
AUROC measures the threshold-independent separability between normal and anomalous pixels.
AP corresponds to the area under the precision–recall curve (AuPRC) and is particularly informative under class imbalance.
FPR\textsubscript{95} denotes the false-positive rate at a fixed 95\% true-positive rate, capturing model reliability in high-recall regimes—an essential requirement for autonomous driving.
For consistency, all metrics follow the evaluation protocol of~\cite{blum2019fishyscapes,chan2021segmentmeifyoucan}.

\subsection{Quantitative Evaluation}
Tables~\ref{tab:results_fs}, \ref{tab:results_smiyc}, and~\ref{tab:results_roadanomaly_only} report results on three widely used anomaly segmentation benchmarks: \textbf{Fishyscapes} (FS-LaF / FS-Static), \textbf{Segment-Me-If-You-Can} (SMIYC-Anomaly / SMIYC-Obstacle), and \textbf{RoadAnomaly}. 
Across all datasets, we focus on two key aspects: (1) comparison with the closest training-free baseline, Maskomaly, and (2) comparison with methods that require re-training, outlier exposure (OE), or additional networks.

\mypara{FS-LaF and FS-Static.}
Relative to Maskomaly, our method yields substantial improvements. 
On \textbf{FS-LaF}, we boost AUROC by \textcolor{green}{+2.86} and AP by \textcolor{green}{+50.21}, while reducing FPR$_{95}$ by \textcolor{green}{24.52}. 
With OOD prompts, AP further rises to \textbf{75.01}, the highest among all methods. 
On \textbf{FS-Static}, our approach improves AP by \textcolor{green}{+12.29} and reduces FPR$_{95}$ by \textcolor{green}{9.22} over Maskomaly. 
Although OE-based methods dominate FS-Static due to its synthetic nature (VOC objects pasted on Cityscapes images), our training-free method remains competitive and outperforms all other non-OE approaches. 
On the more realistic FS-LaF benchmark, our method even surpasses all OE-based methods in AP, demonstrating strong generalization to real-world open-set conditions.

\mypara{SMIYC-Anomaly and SMIYC-Obstacle.}
On \textbf{SMIYC-Anomaly}, our AUPR (92.83) and FPR$_{95}$ (2.89) closely match Maskomaly, indicating comparable performance on diverse, visually salient anomalies. 
However, on \textbf{SMIYC-Obstacle}, which focuses on small, physically realistic road hazards, our method achieves a clear advantage: AUPR improves by \textcolor{green}{+4.68} and FPR$_{95}$ decreases by \textcolor{green}{2.43} over Maskomaly. 
While RPL+CoroCL obtains the best overall results due to OE-based feature learning, our method---requiring neither re-training, OE, nor auxiliary networks---still outperforms many supervised or reconstruction-based baselines and approaches OE performance on SMIYC-Obstacle. 
These results highlight the particular strength of our spatial filtering strategy for fine-grained anomaly localization.

\mypara{RoadAnomaly.}
On the \textbf{RoadAnomaly} benchmark, which contains large and diverse real-world obstacles, our method again improves over Maskomaly in AUROC (\textcolor{green}{+0.13}) and AP (\textcolor{green}{+2.09}), with nearly identical FPR$_{95}$. 
The OOD-prompt variant further enhances both AUROC and AP. 
Although OE-based approaches such as RbA-OE and RPL+CoroCL achieve the strongest overall results, our method significantly outperforms other non-OE baselines (e.g., MSP, MaxLogit, SML) and surpasses several supervised reconstruction/energy-based methods by large margins. 
These findings confirm that our approach generalizes well to real-world anomalous objects without relying on any external OOD training samples.

\begin{table*}[t]
    \centering
    \renewcommand{\arraystretch}{1.2}
    \setlength{\tabcolsep}{6pt}
    \caption{
    Results on FS-LaF and FS-Static validation datasets. Methods are categorized by approach type and whether they require re-training segmentation model (Re-T), outlier exposure (OE), or re-training an extra network (EN). Best results are in \textbf{bold}, second best are \underline{underlined}. "*" indicates reproduced results. "-" indicates unavailable results.
    }
\begin{tabular}{l|c|c|c|c|ccc|ccc}
    \toprule
    \multirow{2}{*}{Method} & \multirow{2}{*}{Category} 
    & \multirow{2}{*}{Re-T} & \multirow{2}{*}{OE} & \multirow{2}{*}{EN}
    & \multicolumn{3}{c|}{FS-LaF} 
    & \multicolumn{3}{c}{FS-Static} \\
    & & & & & AUROC $\uparrow$ & AP $\uparrow$ & FPR$_{95}$ $\downarrow$
      & AUROC $\uparrow$ & AP $\uparrow$ & FPR$_{95}$ $\downarrow$ \\
    \midrule
    
DenseHybrid~\cite{grcic2022densehybrid} 
& Generative& yes & yes & yes 
& \underline{99.01} & 69.79 & 5.09 
& 99.07 & 76.23 & 4.17 \\

Meta-OoD~\cite{chan2021entropy} 
& Confidence (Entropy) & yes & yes & yes  
& 93.06 & 41.31 & 37.69 
& 97.56 & 72.91 & 13.57 \\

PEBAL~\cite{tian2022pixel} 
& Confidence (Energy)  & yes  & yes & no  
& 98.96 &58.81 & \underline{4.76} 
& \underline{99.61} & \underline{92.08} & \underline{1.52} \\

RPL+CoroCL~\cite{liu2023residual} 
& Feature-based & no & yes & yes 
& \textbf{99.39} & 70.61 & \textbf{2.52} 
& \textbf{99.73} & \textbf{92.46} & \textbf{0.85} \\

RbA-OE~\cite{nayal2023rba} 
& Confidence (Mask+Softmax) & yes & yes & no  
& 98.62 & \underline{70.81} & 6.30 
& 98.96 & 75.47 & 3.51 \\

Mask2Anomaly~\cite{rai2023mask2anomaly}   
& Confidence (Mask+Softmax)  & yes & yes  & no 
& - & 69.41 & 9.46 
& - & 95.20 &0.82 \\
\midrule

SynBoost~\cite{di2021pixel}
& Reconstruction& no & no & yes
& 96.21 & 60.58 & 31.02 
& 95.87 & 66.44 & 25.59 \\

MSP~\cite{hendrycks2016baseline}
& Confidence (Softmax)& no  & no  & no  
& 86.99 & 6.02 & 45.63 
& 88.94  & 14.24 & 34.10  \\

MaxLogit~\cite{hendrycks2019scaling}
& Confidence (Logits)& no  & no & no
& 92.00 & 18.77 & 38.13 
& 92.80 & 27.99 & 28.50 \\

SML~\cite{jung2021standardized} 
& Confidence (Logits) & no  & no  & no  
& 96.88 &36.55 &14.53 
& 96.69& 48.67 &16.75 \\

RbA~\cite{nayal2023rba} 
& Confidence (Mask+Softmax) & yes & no & no 
& 96.43 & 60.96 & 10.63 
& 95.26 & 59.14 & 17.71 \\

Maskomaly~\cite{ackermann2023maskomaly} 
& Confidence (Mask+Softmax) & no  & no  & no 
& 94.45*  & 17.20*  & 35.72*  
& 94.13* & 53.06* & 40.85* \\
\midrule

\textbf{Ours} 
& Confidence + Spatial Filter & no & no & no 
& 97.31 & 67.41 & 11.20 
& 95.08 & 65.35 & 31.63 \\

\textcolor{gray}{\textbf{Delta (vs.~Maskomaly)}} 
& & & & 
& {\color{green}{+2.86}} & {\color{green}{+50.21}} & {\color{green}{-24.52}}
& {\color{green}{+0.95}} & {\color{green}{+12.29}} & {\color{green}{-9.22}} \\

\textbf{Ours + OOD prompts} 
& Confidence + Spatial Filter & no & no & no 
& 97.75 & \textbf{75.01} & 11.20 
& 95.08 & 65.35 & 31.63 \\

\textcolor{gray}{\textbf{Delta (vs.~Maskomaly)}} 
& & & & 
& {\color{green}{+3.30}} & {\color{green}{+57.81}} & {\color{green}{-24.52}}
& {\color{green}{+0.95}} & {\color{green}{+12.29}} & {\color{green}{-9.22}} \\
\bottomrule
\end{tabular}
    \label{tab:results_fs}
\end{table*}

\begin{table*}[t]
    \centering
    \renewcommand{\arraystretch}{1.2}
    \setlength{\tabcolsep}{6pt}
    \caption{
    Results on SMIYC-Anomaly and SMIYC-Obstacle validation dataset. Methods are categorized by approach type and whether they require re-training segmentation model (Re-T), outlier exposure (OE), or re-training an extra network (EN). 
    Best results are in \textbf{bold}, second best are \underline{underlined}. "*" marks reproduced results. "-" indicates unavailable results.
    }
\begin{tabular}{l|c|c|c|c|cc|cc}
    \toprule
    \multirow{2}{*}{Method} & \multirow{2}{*}{Category} 
    & \multirow{2}{*}{Re-T} & \multirow{2}{*}{OE} 
    & \multirow{2}{*}{EN}
    & \multicolumn{2}{c|}{SMIYC-Anomaly} 
    & \multicolumn{2}{c}{SMIYC-Obstacle } \\
    & & & & & AUPR $\uparrow$ & FPR$_{95}$ $\downarrow$ 
      & AUPR $\uparrow$ & FPR$_{95}$ $\downarrow$  \\
    \midrule

DenseHybrid~\cite{grcic2022densehybrid} 
& Generative & yes & yes & yes 
& 61.08 & 52.65 
& 89.49 & 0.71 \\

Meta-OoD~\cite{chan2021entropy} 
& Confidence (Entropy) & yes & yes & yes 
& {80.13} & {17.43} 
& \underline{94.14} & \underline{0.41} \\

PEBAL~\cite{tian2022pixel} 
& Confidence (Energy) & yes & yes & no 
& 53.10 & 36.74 
& 10.45 & 7.92 \\

RPL+CoroCL~\cite{liu2023residual} 
& Feature-based & no & yes & yes 
& {88.55} & {7.18} 
& \textbf{96.91} & \textbf{0.09} \\
\midrule

Mahalanobis~\cite{lee2018simple} 
& Feature-based  & no & no & no 
& 22.50 & 86.40 
& 25.90 & 26.10 \\

SynBoost~\cite{di2021pixel} 
& Reconstruction & no & no & yes 
& 68.80 & 30.90 
& 81.40 & 2.80 \\

MSP~\cite{hendrycks2016baseline} 
& Confidence (Softmax) & no & no & no 
& 40.40 & 60.20 
& 43.40 & 3.80 \\

SML~\cite{jung2021standardized} 
& Confidence (Logits) & no & no & no 
& 21.68 & 84.13 
& 18.60 & 91.31 \\

Maskomaly~\cite{ackermann2023maskomaly} 
& Confidence (Mask+Softmax) & no & no & no 
& \textbf{94.13}* & \textbf{2.83}* 
& 88.47* & 2.84* \\
\midrule

\textbf{Ours} 
& Confidence + Spatial Filter & no & no & no 
& 92.83 & \underline{2.89} 
& 93.15 & \underline{0.41} \\

\textcolor{gray}{\textbf{Delta (vs. Maskomaly)}} 
& & & & 
& \textcolor{red}{-1.30} & \textcolor{red}{+0.06} 
& \textcolor{green}{+4.68} & \textcolor{green}{-2.43} \\

\textbf{Ours + OOD prompts} 
& Confidence + Spatial Filter & no & no & no 
& \underline{92.87} & \underline{2.89}  
& 93.15 & \underline{0.41} \\

\textcolor{gray}{\textbf{Delta (vs. Maskomaly)}} 
& & & & 
& \textcolor{red}{-1.26} & \textcolor{red}{+0.06} 
& \textcolor{green}{+4.68} & \textcolor{green}{-2.43} \\

\bottomrule
\end{tabular}
    \label{tab:results_smiyc}
\end{table*}

\begin{table*}[t]
    \centering
    \renewcommand{\arraystretch}{1.2}
    \setlength{\tabcolsep}{6pt}
    \caption{
    Results on RoadAnomaly dataset. Methods are categorized by approach type and whether they require re-training segmentation model (Re-T), outlier exposure (OE), or re-training an extra network (EN). The best results are shown in \textbf{bold}, and the second best are \underline{underlined}. "*" indicates reproduced results. "-" indicates unavailable results.
    }
\begin{tabular}{l|c|c|c|c|ccc}
    \toprule
    \multirow{2}{*}{Method} & \multirow{2}{*}{Category} 
    & \multirow{2}{*}{Re-T} & \multirow{2}{*}{OE} 
    & \multirow{2}{*}{EN} 
    & \multicolumn{3}{c}{RoadAnomaly} \\
    & & & & & AUROC $\uparrow$ & AP $\uparrow$ & FPR$_{95}$ $\downarrow$ \\
    \midrule

DenseHybrid~\cite{grcic2022densehybrid} 
& Generative& yes & yes & yes 
& - & 63.97 & 43.20 \\

PEBAL~\cite{tian2022pixel} 
& Confidence (Energy) & yes & yes & no 
& 87.63 & 45.10 & 44.58 \\

RPL+CoroCL~\cite{liu2023residual} 
& Feature-based & no & yes & yes 
& \underline{95.72} & 71.61 & 17.74 \\

RbA-OE~\cite{nayal2023rba} 
& Confidence (Mask+Softmax) & yes & yes & no 
& \textbf{97.99} & \textbf{85.42} & \textbf{6.92} \\

Mask2Anomaly~\cite{rai2023mask2anomaly} 
& Confidence (Mask+Softmax) & yes & yes & no 
& - & \underline{79.70} & \underline{13.45}  \\

Image Resyn.~\cite{lis2019detecting}
& Reconstruction& no & no & yes 
& -& 76.40 & 48.10  \\

SynBoost~\cite{di2021pixel}
& Reconstruction& no & no & yes 
& 85.23 & 41.83 & 59.72 \\
 \midrule

MSP~\cite{hendrycks2016baseline}
& Confidence (Softmax)& no & no & no 
& 73.76 & 20.59 & 68.44 \\

MaxLogit~\cite{hendrycks2019scaling}
& Confidence (Logits)& no & no & no 
& 77.97 & 24.44 & 64.85 \\

SML~\cite{jung2021standardized} 
& Confidence (Logits) & no & no & no 
& 81.96 & 25.82 & 49.74 \\

RbA~\cite{nayal2023rba} 
& Confidence (Mask+Softmax) & yes & no & no 
& 95.60 & 78.45 & 11.83 \\

Maskomaly~\cite{ackermann2023maskomaly} 
& Confidence (Mask+Softmax) & no & no & no 
& 94.89* & 73.19* & 28.87* \\
\midrule

\textbf{Ours} 
& Confidence + Spatial Filter & no & no & no 
& 95.02 & 75.28 & 28.89 \\

\textcolor{gray}{\textbf{Delta (vs.~Maskomaly)}} 
& & & & 
& {\color{green}{+0.13}} & {\color{green}{+2.09}} & {\color{red}{+0.02}} \\

\textbf{Ours + OOD prompts}
& Confidence + Spatial Filter & no & no & no 
& 95.03 & 75.56 & 28.89 \\

\textcolor{gray}{\textbf{Delta (vs.~Maskomaly)}} 
& & & & 
& {\color{green}{+0.14}} & {\color{green}{+2.37}} & {\color{red}{+0.02}} \\

\bottomrule
\end{tabular}
    \label{tab:results_roadanomaly_only}
\end{table*}

\subsection{Qualitative Evaluation}
\begin{figure*}[t!]
\centering
\includegraphics[trim=0 0 0 0,clip,width=1\linewidth]{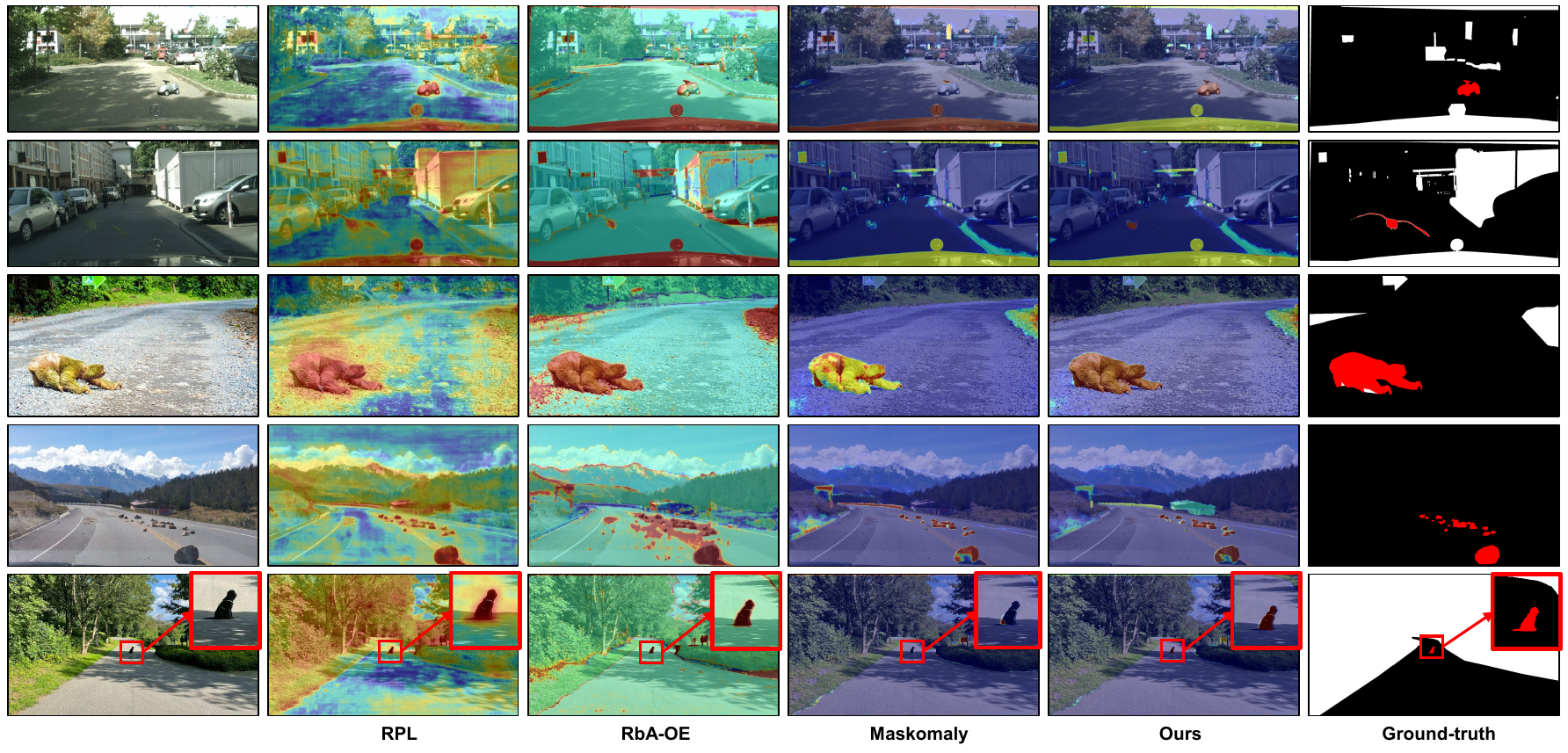}
\vspace{-7mm}
\caption{
Qualitative comparison of anomaly segmentation performance across various state-of-the-art (SOTA) methods and benchmark datasets.
From top to bottom: samples are selected from FS-LostAndFound, FS-Static, SMIYC-Anomaly, SMIYC-Obstacle and RoadAnomaly datasets.
From left to right: input RGB image, followed by predictions from RPL, RbA-OE, Maskomaly, our proposed method and the Ground-truth.
Each result displays an anomaly heatmap overlaid on the original image to highlight anomalous regions—red indicates high anomaly scores (abnormal areas), while blue indicates low anomaly scores (normal areas). In the Ground-truth, pixels colored red represent OOD objects, black pixels denote ID objects, and white pixels indicate regions to be ignored. Zoom in for better visualization.
}
\label{fig:qualitative_results}
\end{figure*}

Fig.~\ref{fig:qualitative_results} presents qualitative anomaly segmentation results from RPL, RbA-OE, Maskomaly, and our proposed method. For clearer visualization, each anomaly map is overlaid on the corresponding RGB image, with colors ranging from blue to red indicating increasing anomaly scores. As shown, our method not only reduces false positives (FP) but also significantly mitigates false negatives (FN), resulting in more accurate detections of small anomalous objects on the road.

\section{Conclusion}
\label{sec:conclusion}

We proposed a lightweight, road-aware anomaly segmentation framework for autonomous driving. 
Built on a frozen Mask2Former model, our method integrates a road-aware filtering module with acceptance- and rejection-based soft masks to produce robust anomaly maps without retraining or outlier exposure. 
The road-aware module derives polygonal road priors from road-related queries, detects gap regions as anomaly candidates, and refines them using CLIP-based zero-shot classification to suppress semantically plausible false positives. 
By combining spatial priors, semantic cues, and uncertainty-aware fusion, the proposed approach achieves accurate and interpretable anomaly segmentation. 
Experiments on three real-world benchmarks show consistent improvements over training-free baselines and competitive performance against retraining-based methods, reaching the best average precision on Fishyscapes LostAndFound. 
Overall, the proposed post-processing pipeline provides an efficient and practical solution for open-world anomaly detection when retraining is infeasible.

\section{Acknowledgment}
The authors would like to thank the Federal Ministry for Economic Affairs and Climate Action for supporting this work under the project “Gaia-X 4 AMS”.



\bibliography{ref}
\bibliographystyle{IEEEtran}

\end{document}